# SUPERPOSITION AS DATA AUGMENTATION USING LSTM AND HMM IN SMALL TRAINING SETS

Akilesh Sivaswamy, Evgeniy Pavlovskiy
{a.sivasvami, e.pavlovskiy}@g.nsu.ru
Stream Data Analytics and Machine Learning lab
Novosibirsk State University

**Abstract**—Considering audio and image data as having quantum nature (data are represented by density matrices), we achieved better results on training architectures such as 3-layer stacked LSTM and HMM by mixing training samples using superposition augmentation and compared with plain default training and mix-up augmentation. This augmentation technique originates from the mix-up approach but provides more solid theoretical reasoning based on quantum properties. We achieved 3% improvement (from 68% to 71%) by using 38% lesser number of training samples in Russian audio-digits recognition task and 7,16% better accuracy than mix-up augmentation by training only 500 samples using HMM on the same task. Also, we achieved 1.1% better accuracy than mix-up on first 900 samples in MNIST using 3-layer stacked LSTM.

## 1. INTRODUCTION

One of the main challenges in deep learning is training on small training sets. A good regularization and augmentation technique known as *mix-up* [1], was introduced in ICLR-2018. We revised this technique based on quantum information and developed a quantum superposition technique for data augmentation. The technique consists of: (1) considering data in terms of quantum information either considering some embeddings by deep architectures; (2) usage of density matrices as intermediate representations of samples; (3) usage of the superposed sample matrices. All the experiments we conducted using this technique require increasing the data set size by introducing new transformed and meaningful samples obtained from original samples to provide the network with more diversity and exposure to learning data. This was reported in a scientific work, wherein the researchers concluded that experimental results run using 4-fold cross-validation and reported in terms of Top-1 and Top-5 accuracy, indicate that cropping in geometric augmentation significantly increases CNN task performance [2].

The main goal of this work is to study and establish results obtained from implementing the superposition principle in HMM and LSTM for speech classification task on a small audio training set. The importance of research direction along Russian speech recognition and HMM deployment in this direction, especially in noisy conditions, has been explained in [3]. As Russian pronunciation varies greatly among different groups of people, by experiment and comparison, scientists found that Russian recognition results are not too satisfactory. Hence, more efficient recognition training algorithms are needed for practical purpose [4].

We used the Hidden Markov Model (HMM) as it is one of most popular tools used in speech recognition from the 80s. Even today, it is used to solve sequence analysis problems. In early days, this algorithm was used to code in noisy digital communication links, thus proving to be a good choice for speech classification tasks as in our case as it calculates output probability of speech parameters to the HMM model using the probability density function, searches the best state sequence and obtains the recognition result using the criteria of maximum posteriori probability [5].

The LSTM model makes use of an activation function just like a Convolution Neural Network, except that this kind of network can work on sequential data, as it also considers the previously received inputs. It is an 'upgrade' of the Recurrent Neural Network (RNN) in the way that it deals with vanishing and exploding gradient problem, having proved its improved speech recognition accuracy for the context independent 126 output state model, context dependent 2000 output state embedded size model [6].

The Long-Short Term Memory (LSTM) is much faster and more accurate than both standard Recurrent Neural Nets (RNNs) and time-windowed Multilayer Perceptrons (MLPs) [7]. Researchers also show that a two-layer deep LSTM RNN where each LSTM layer has a linear recurrent projection layer can exceed state-of-the-art speech recognition performance [8]. Henceforth, we stand strong in our decision to use a 3-layer stacked LSTM to train.

## 2. EXPERIMENT SETUP

The experiments were run on two different datasets with small number of training samples, as well as with small number of classes: (a) Russian audio-digits dataset (1775 samples, digits spelled by Russian students from 0 to 9), (b) MNIST. The following architectures were tested: (i) 3-layer stacked LSTM, (ii) HMM (used on audio only).

We superpose initial samples using the formula:

$$\lambda^2 D_i^2 + (1 - \lambda^2)D_j^2 + \lambda * \sqrt{(1-\lambda^2)}(D_i * D_j + D_j * D_i) \qquad (1)$$

where, $D_i$ and $D_j$ are the Density Matrices obtained from the *i*th and *j*th samples in the dataset, by multiplying the conjugate of the state vector with the state vector itself. The state vector used here was obtained from the dense layer before the final dense layer. The third term in the above equation is the interference. This approach of using density matrices was only applied to audio in LSTM architecture and it involves training the model on one Neural Network (LSTM) and predicting all the samples from the model to form Density Matrices (multiplying layer embeddings from the second-last dense layer by its transpose to get 64 x 64 matrix for each sample) and finally by using superposition with interference, we train a new model with the same 3-layer LSTM architecture.

In the case of MNIST, the same approach was also used in LSTM architecture, but the density matrices were replaced by sample matrices:

$$\lambda^2 S_i^2 + (1 - \lambda^2)S_j^2 + \lambda * \sqrt{(1-\lambda^2)}(S_i * S_j + S_j * S_i) \qquad (2)$$

We also entitled ourselves to another approach wherein the sample matrices themselves were used instead of density matrices, but without the interference term:

$$\lambda S_i^2 + \sqrt{(1-\lambda^2)}S_j^2 \qquad (3)$$

where, $S_i$ and $S_j$ are the sample matrices obtained from the *i*th and *j*th samples in the dataset. In both cases, the $\lambda^2$ values taken were: 1, 0.2, 0.5 and 0.8. The above approach is the 'quantum version' of mix-up or superposition without interference as it uses quantum probabilities.

However, in mix-up approach, the following formula was used:

$$\lambda S_i + (1 - \lambda)S_j \qquad (4)$$

where, the λ values taken were: 1, 0.2, 0.5 and 0.8. All the above approaches were implemented in Keras framework using TensorFlow backend with GPU support and built-in categorical cross entropy loss function. Adam optimizer was used as even though Adam shows marginal improvement over SGD with momentum, it adapts learning rate scale for different layers instead of hand picking manually as in SGD [9]. Note that we used only the first 1100 samples of audio to train as the memory usage was limited and for samples larger than 1100, we many a times experienced memory errors and long delays. Since we required memory to perform experiments, we tried with the maximum possible number of samples: 1100, beyond which the system throws a memory error.

In HMM, we did not use the Density Matrix approach as the fit function used for training the HMM model accepts only 2D inputs and it is not possible to flatten a Density Matrix to use for training as GPU was not supported to train very large vectors. This is because the built-in functions used in HMM architecture did not utilize TensorFlow backend or Keras framework.

The target labels were one-hot encoded. These were represented as soft labels. For instance, an example could be of the form: [0, 0.2, 0, 0, 0, 0, 0, 0, 0, 0.8]. This example shows a case where samples from class '1' and class '9' were selected for mixing or superposing.

## 3. RESULTS

The first result was obtained from training on first 900 samples of MNIST with mix-up or superposition on LSTM:

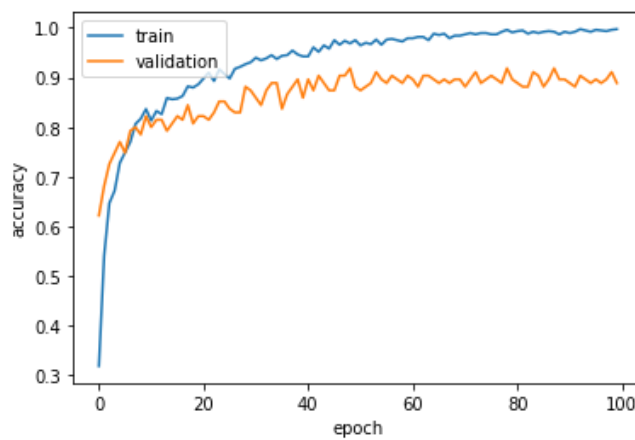

**Figure 1. Model accuracy on MNIST trained with default setting on first 900 samples.**

From the above graph, we can see that the model overfits slightly as the difference between the training and validation accuracy diverges from 10 epochs of training. This showcases the most common problem of training on small data sets: overfitting. Even though the training accuracy almost reaches 100%, we can see that the validation accuracy only manages to touch 90%. The test accuracy obtained was 90.25% without mix-up and superposition.

We noted that the mix-up approach on just 900 samples of MNIST significantly improved the performance by 3% on the original test dataset, in comparison to the 90.25% produced by plain training.

For superposition of MNIST with interference, we plotted the graphs with validation curves to study the behavior of the model during training. The results that we obtained on original test data was much better (1% increase) than that of the mix-up approach.

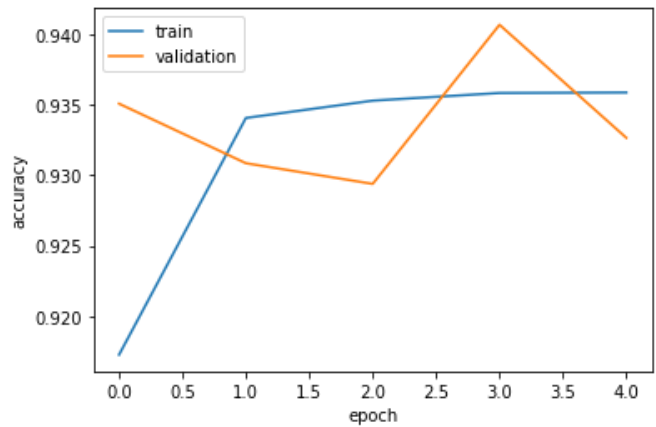

**Figure 2. Model accuracy on MNIST with superposition and interference on first 900 samples.**

The same LSTM architecture was tested on our audio dataset, where our model overfitted on the training data to a considerable extent as shown below:

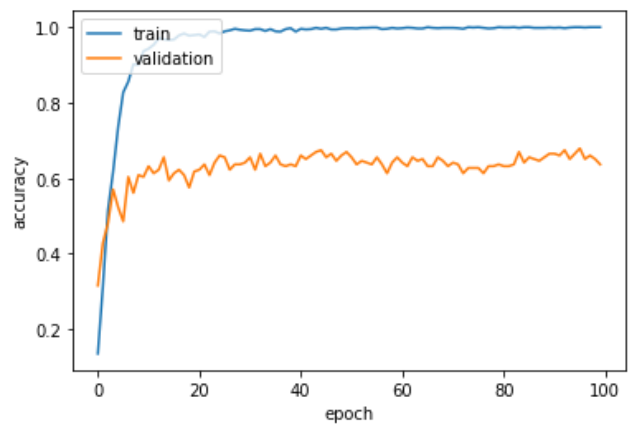

**Figure 3. Model Accuracy on Audio with default setting.**

The idea behind overfitting is that the training accuracy differs from the validation accuracy to a significant extent. This is clearly visible from the graph. The model achieved a prediction accuracy of 68.88% on the test set split with overfitting.

The mix-up approach produced very poor results (10% test accuracy). The results obtained from our Density Matrix approach was also unsatisfactory as only 35% accuracy was produced on the original audio test set of 530 samples train-test split.

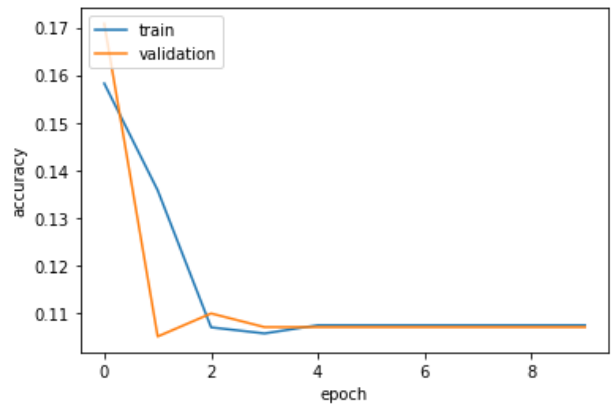

**Figure 4. Model accuracy on Audio using Density Matrix approach.**

However, we achieved good results by training on first 1100 superposed audio samples with interference from scratch without using any previous layer embeddings:

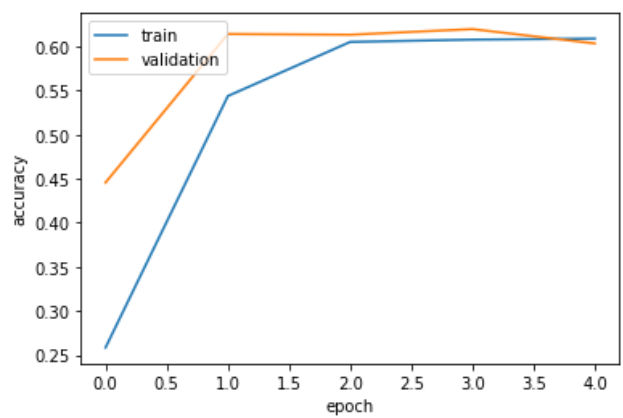

**Figure 5. Model accuracy on Audio with superposition and interference on first 1100 samples.**

  Even though this superposition with interference model only achieved around 60% accuracy and almost the same validation accuracy without overfitting, it managed to achieve 71.38% on unseen test split of 0.2 on the entire audio dataset (353 files) by training only on 1100 superposed data with interference. Another important observation that we made from our results is that this model predicts more accurately or does lesser mistakes in predicting 'apparent pure states' or intra class superposition i.e. the models is able predict correctly on 2 different samples with the same labels. In such a case, the superposition of sample with label '0' and another sample with the same label will have a one-hot target [1.0, 0, 0, 0, 0, 0, 0, 0, 0, 0]. The model seems to predict better on intraclass superposed states than on interclass superposed states. The results obtained were satisfactory and decent for a model trained on superposed states with interference obtained only from 1100 samples from the original audio dataset.

  Lastly, we deployed the HMM model for training on audio with superposition ('quantum version of mix-up or without interference), with mix-up approach and without superposition and mix-up. Here, we did not use Density Matrix approach as the fit function used for training the HMM model accepts only 2D inputs and it is not possible to flatten a Density Matrix to use for training as GPU was not supported to train very large vectors. Since we were only able to use 500 samples (50 samples from each class) to train for mix-up and superposition, we established a baseline score for just 500 samples of 28.39% without mix-up and superposition.

  We then used the mix-up approach on 500 samples, with the same mix proportion of lambda values used in LSTM, but here we obtained an accuracy of 29.5%. But, when using superposition in HMM, we saw a great increase in test accuracy (36.66%) of about 8.3% from the baseline score.

  During training of HMM, we only performed superposition on intra class superposed samples or 'apparent pure states' or in other words, our superposed training dataset only had the original samples (without superposition or pure states) and apparent pure states. We did not use inter class superposed samples because firstly, we did not have enough memory to run the code with a very large superpose dataset and secondly, our observation during the previous experiment using LSTM on superposed data showed that the model predicted better on intra class superposition.

## 4. RESULTS SUMMARY

| No. | Dataset | Number of Samples | Augmentation | Train accuracy | Test accuracy on original data split |
|---|---|---|---|---|---|
| 1 | MNIST | 900 | – | 66% | 68% |
| 2 | MNIST | 900 | mix-up | 93.5% | 93% |
| 3 | **MNIST** | **900** | **superposition** | **94%** | **94.1%** |
| 4 | Audio Digits | 1775 | – | 100% | 68% |
| 5 | Audio Digits | 1100 | Density Matrix | 10% | 35% |
| 6 | Audio Digits | 1100 | mix-up | 30% | 10% |
| 7 | **Audio Digits** | **1100** | **superposition and interference** | **84%** | **71%** |
| 8 | Audio Digits | 1775 | – | – | 40.96% |
| 9 | Audio Digits | 500 | – | – | 28.39% |
| 10 | Audio Digits | 500 | mix-up | – | 29.50% |
| 11 | **Audio Digits** | **500** | **superposition** | – | **36.66%** |

## 5. CONCLUSIONS

An accuracy of 71.38% was achieved on audio original test split by implementing superposition with interference on only 1100 samples from the original 1775 samples. The model overfits when we train without superposition. The superposition principle also works on MNIST dataset with 900 samples and outperforms mix-up approach and default test accuracy by producing 94.1% on test dataset with original test dataset with 10,000 samples. Mix-up approach did not work with audio in LSTM architecture. The approach of applying superposition on density matrices obtained from previous layer embeddings was not good and produced poor results on audio with LSTM architecture. The implementation of superposition principle on HMM was successful as it achieved 36.66% accuracy. Superposition acts as a good regularization technique when we have a small training set.

## ACKNOWLEDGEMENTS
The reported study was funded by RFBR according to the research project № 19-29-01103.

## REFERENCES

[1] Zhang, H., Cisse, M., Dauphin, Y.N., Lopez-Paz, D.: mixup: Beyond empirical risk minimization. In: Proceedings of the International Conference on Learning Representations (2018).

[2] Taylor L, Nitschke G. Improving deep learning using generic data augmentation. arXiv preprint arXiv:1708.06020. 2017 Aug 20.

[3] Abdul Ahad, Ahsan Fayyaz, Tariq Mehmood, "Speech recognition using Multilayer Perceptron", IEEE con., 2002.

[4] Vabishchevich P. Exploration of automatic speech recognition for Russian in noisy conditions. State University of New York at Binghamton; 2015.

[5] Jia Zheng and Zhi-Qiang Liu, "Type 2 Fuzzy Hidden Markov Models and their application to Speech Recognition", IEEE TRANSACTIONS ON FUZZY SYSTEMS, VOL. 14, NO. 3, JUNE 2006.

[6] Sak H, Senior A, Beaufays F. Long short-term memory based recurrent neural network architectures for large vocabulary speech recognition. arXiv preprint arXiv:1402.1128. 2014 Feb 5.

[7] Graves A, Schmidhuber J. Framewise phoneme classification with bidirectional LSTM and other neural network architectures. Neural Networks. 2005 Jul 1;18(5-6):602-10.

[8] Sak H, Senior A, Beaufays F. Long short-term memory recurrent neural network architectures for large scale acoustic modeling. In Fifteenth annual conference of the international speech communication association 2014.

[9] Kingma DP, Ba J. Adam: A method for stochastic optimization. arXiv preprint arXiv:1412.6980. 2014 Dec 22.